\documentclass{llncs}

\def \Rmath{\mbox{$I\hspace{-1.3mm}R$}}
\def \Nmath{\mbox{$I\hspace{-1mm}N$}}
\def \Zmath {\mbox{$Z\hspace{-1.7mm}Z$}}

\title{{\bf Reasoning with Intervals on Granules}}
\author{
Sylviane R. Schwer\thanks{This work is partially supported by the 
MENRT, throught REANIMATIC project}
}
\institute{L.I.P.N. UPRESA CNRS 7030\\
Institut Galil\'ee, Universit\'e Paris 13\\
99, Bd J-B Cl\'ement\\
Fr. 93430 Villetaneuse\\
e-mail: schwer@lipn.univ-paris13.fr
}

\begin{document}
\pagestyle{empty}
\maketitle
\begin{abstract}
The formalizations of periods of time inside a linear model of Time 
are usually based on the notion of intervals, that
may contain or may not their endpoints. This is not enought when the 
periods are written in terms
of  coarse granularities with respect to the event taken into 
account. For instance, how to express the inter-war
period in terms of a {\em years} interval?

This paper presents a new type of intervals, neither open ,nor closed 
or open-closed and the extension of operations on intervals of this 
new type,
in order to reduce the gap between the discourse related to temporal 
relationship and its translation
into a discretized model of Time.

\end{abstract}

\section{Introduction}
As Humberstone mentionned it in its introduction \cite{Hum79}, ever 
since Zeno, philosophers have  been aware that there is something 
problematic about the notion of an instant,
  or moment of time. The problem of the instant of change is the 
following: is an object which changes in a certain
respect at a certain moment in its pre-change or its post-change 
state, at that moment ? If the answer is ``neither'', it
violates the law of excluded middle. If the answer is ``both'', it 
violates the law of non-contradiction.
In order to get a more adequate answer, the notion of interval of 
time, or period, has been prefered,  a period being an uninterrupted 
stretch of time,
during which something may happen. This notion of period is an entity 
in its own right and not a set of points. This response
can be traced back to Aristotle and even to the Stoicism.

In european languages, one refers to time as instants as well as periods,
  related to calendar units and clock units (e.g. {\it  years}, {\it 
weeks}, {\it months},
{\it  hours} and {\it seconds}).  For instance, in English, ``We met 
them yesterday'' refers to a period of time; but
``We met them at eleven'' refers to an instant, introduced by {\em 
at}. In German for days or part of days, the preposition {\em am}
is used, and for longer periods, {\em im} is used: one says ``Am 
Dienstag'' but ``Im Sommer''.

Depending on the context, a same unit is viewed as a point or an interval.
Allen and Hayes \cite{AlH87} have proposed a model that takes into 
account the two objects. In order to avoid confusion
  between mathematical objects and temporal object, they propose 
periods instead of intervals and moments as points,
following Humberstone \cite{Hum79}.

Most human activities are linked to the social
artefacts that are named calendars and refer to calendar units.
These units are also called granularities \cite{Bet98} or 
chronologies \cite{Sch2002} or time units \cite{WJS93},
and this research domain has been recognized to be an important issue.
Conversions are proposed when two granularities are commensurable, as
  {\it months} and {\it  years} are.
The method is always the same. If a
granularity $f$, say {\it months}, is finer than a granularity $g$, 
say {\it years},
an occurrence of $f$, a  {\it  month}, is translated into the 
occurrence of $g$,
the {\it year}, that contains it: $June2000$ is so converted into $2000$.
  In the other way,
an occurrence of $g$ is converted into the interval of all 
occurrences of $f$ that are contained in it:  $2002$ is
converted into $[January2002, \ December2002]$. Conversions between 
incommensurable granularities are
proposed also when these two granularities have in commun a finer 
granularity, {\it Days} with respect to {\it  months} and {\it weeks}.

In \cite{Euz95,BWJ96}  has been pointed out
that a constraint about a temporal relationship in one granularity 
may not be preserved in another granularity.
As an example, if a constraint
says that an event must happen in the {\it day} that follows the {\it 
day} when another event happens,
then this constraint cannot be
translated into one in terms of {\it  hours} because it is incorrect 
to say that the second event
must happen within $24$ hours after the first event happens.
The reader can see that the solution is $x$ hours for some $x$ such 
that $1 \leq x \leq 47$.
  This constraint cannot be also translated automatically into one in 
terms of {\it months} because
the first event may
occur the last day of a month, the next day taking then place in an 
other month.

In Databases framework, the same problem has been risen \cite{DyS93}. 
In fact, it is clear that, when the Time line,
supposed to be continuous, is partitioned into intervals with 
non-null length, called granules of time,
  any instant is approximated with such a granule. Exactly in the same 
manner as any measurement is an
  approximation of what is measured. Two instants that are located 
within the same granule, inside a granularity,
may be strictly ordered inside a finer granularity. Different 
approaches have been proposed, for dealing with that problem
  of precision or indeterminacy,  based on fuzzy sets or possibilistic 
distributions \cite{DuP91}.
Much attention has been paid about the conversion of temporal 
expression from one granularity to another, about
an information which is supposed to be timestamped with the good 
granularity according to its management.
But until now, no attention has been paid to the discrepancy between 
the time granularity expressed in the discourse and
the time granularity induced by the knowledge level, that induces 
temporal relationships on finer granularity that
are contained in the knowledge level, but not expressed in the discourse.

  For instance, the worldwide II war is
called 1939-1945 war, which express the fact that this war began in 
the course of the year 1939 and ended in
the course of year 1945. The same thing can be said about 1914-1918 war.
And the period between these two wars
is obviously, for a human being, the period 1918-1939, which is 
understood by every one to begin in the course of
1918 and end in the course of 1939.

This problem is concerned with how to express indeterminacy related 
to the expression {\it in the course of}
which can be expressed or simply inferred by the knowledge of he context.

  That is the problem we address in
this paper which can be divided in three steps: {\it i.e.\ } (1) how 
to express the difference between  ends of intervals that are wholly 
included
in the validity period of the fact taken into account and the 
endpoints of intervals that are partially included in it, (2)
   how to manage with it inside a
granularity and (3) how to go from one granularity to another one.

The paper is organized as follows.
We begin with two examples that motivate this work. The first one is 
taken from a book about french history, the
second one is inspired by the REANIMATIC project, which is devoted to 
a medical datawarehouse.
Then, refering to Allen's work and to the concept of chronology and 
calendar, we propose a new kind of interval,
buit on granules, with three types of ends. In forth part, we recall 
the boolean calculus on usual intervals (with two types of
ends) and  extend the calculus to intervals on granules. We end 
showing how these new intervals and their calculus allow to
solve our two example problems.

  \section{Two examples}
The first example concerns the linguistic temporal locution {\em 
entre-deux-guerres},  that we want to translate inside the framework 
of a
  formal model. The second one is the description of
how beds are managed in a french hospital.
\subsection{The inter-war period}
In the dictionary \cite{Har78}, one can read ``entre-deux-guerres : 
The inter-war period (1918-1939)''. Following
this notation, the first worldwide war period is (1914-1918) and the 
second worldwide war period is (1939-1945).
Everybody understands that these three periods are adjacent, or 
following Allen's vocabulary
\cite{AlH85}, (1914-1918) {\em meets} (1918-1939) and (1918-1939) 
{\em meets} (1939-1945).
It would be a pity not to be
able to maintain this knowledge when storing it in a database, 
without the precise date (with the granularity
  {\it days}).

Temporal elements provided by usual Data Models for Time
are either points, intervals or subsets of a linear and discrete set, 
such as the set of positive integers \Nmath,
  or as the set of integers \Zmath.
But points, in Data Model for Time, are not without duration. They 
represents an occurrence of a time unit,
which is an uninterrupted stretch of time with duration. So that an 
interval of such objects is in fact an interval
of intervals.
There are only four types of intervals in \Nmath and \Zmath:
the open interval, in which the two endpoints are outside the interval;
the closed interval, in which the two endpoints are inside the interval;
the open-closed interval, in which the left endpoint is  outside  and 
the right endpoint is inside the interval;
the closed-open interval, in which the left endpoint is  inside and 
the right endpoint is outside the interval.
Let us take all occurrences of the unit {\it years} as points and 
apply on the first world-war period (1914-1918) the four types of 
intervals.
The closed interval  [1914, 1918]  says that the first world-war 
began at the begining of the year 1914 and ended at the end of the 
year 1918;
the open interval  ]1914,1918[ says that the first world-war began at 
the begining of the year 1915 and ended at the end of the year 1917;
  the closed-open interval [1914, 1918[ says that the first world-war 
began at the begining of the year 1914 and ended at the end of the 
year 1917;
and the open-closed interval  ]1914, 1918] says that the first 
world-war began at the begining of the year 1915 and ended at the end 
of the year 1918.
   None of these four intervals models the qualitative meaning of the 
linguistic locution {\it  in the course of}, semantically linked to 
the expression (1914-1918) as
shown in Figure \ref{war}.

\begin{figure}
\setlength{\unitlength}{6mm}
\caption{\label{war}(1914,1918)}
\begin{center}
\begin{picture}(20,1.5)
\put(2,1){\line(1,0){17}}
\put(0,0){\line(1,0){20}}
\multiput(0,0)(4,0){6}{\line(0,1){1.1}}
\put(1.5,0.2){1914}
\put(5.5,0.2){1915}
\put(9.5,0.2){1916}
\put(13.5,0.2){1917}
\put(17.5,0.2){1918}
\end{picture}
\end{center}
\end{figure}
In order to have a good representation of this kind of problem, we 
need a new kind of object which is both sharable as any
uninterrupted stretch of time and irresolvable as atomic object. Let 
us call this kind of object a {\it granule}.

\subsection{Beds managment in hospitals}
In a french hospital, a day begins at 8 a.m.\ and lasts 24 hours. If 
Jack goes to the hospital at 2 p.m.\ on monday
march  $13th$ and leaves at 10 a.m.\ on friday march  $17th$, he will 
be registered and will be
requested to pay for 5 days. If George goes to the same hospital at 6 
p.m.\ on monday
march $6th$ and leaves at 10 a.m.\ on monday march  $13th$, he will 
be registered and will be
requested to pay for 8 days.  If Karl goes to the same hospital from 
11 a.m.\ to 4 p.m.\ on  monday march
  $13th$,  he will be recorded and will be requested to pay for 1 day.

  Suppose that they occupy the same bed,
let us say bed $13$. The registration database is described in Table 
\ref{Tab1}.

\begin{table}
\begin{center}
\caption{\label{Tab1}The relational table of staying days  in the hospital}
\begin{tabular}{|c|c|c|c|c|}
\hline
Bed & Name & admission-day & exit-day \\ \hline \hline
13 & Jack & 03/13 & 03/17 \\ \hline
13 & George & 03/06 & 03/13\\ \hline
13 &Karl & 03/13 & 03/13 \\ \hline
\end{tabular}
\end{center}
\end{table}

  The amounts of chage dues by the patients are computed following 
Table \ref{Tab2}, all intervals are supposed to be closed.

The stay includes the two endpoints days, even if they are partial. 
That so, the day 03/13 is paid three times
instead of one time, since
according to the hospital database, for the same bed, 5+8+1=14 days 
have been accounted from 03/06 until
03/17 (that is for 12 days) thought at any moment there is at most 
one person in the bed.
The solution that consists in changing the time unit to hours
would  induce perhaps less liberty for a patient to leave the 
hospital and a change
inside the database of the hospital.
\begin{table}
\begin{center}
\caption{\label{Tab2}The closed interval's solution: the usual case}
\vspace{2pt}
\begin{tabular}{|c|c|c|c|}\hline
Bed & Name & stay [x,y] & days due=y-x+1 \\ \hline \hline
13 & Jack &[ 03/13, 03/17] & 5\\ \hline
13 & George & [03/06,  03/13] & 8\\ \hline
13 &Karl & [ 03/13, 03/13] & 1\\ \hline
\end{tabular}
\end{center}
\end{table}
Let us show  that none of  closed, open, open-closed or closed-open intervals,
with the same endpoints, are able to model this reality,
  that is to make the
social security to pay only one day per bed, when a bed is occupied 
(partly or not) this day without to change the
  granularity. {\it Days due} computes the number of days inside the interval.

The open interval will give the following table \ref{Tab3}:
\begin{table}
\begin{center}
\caption{\label{Tab3}the open interval's solution}
\vspace{2pt}
\begin{tabular}{|c|c|c|c|}\hline
Bed & Name & stay ]x,y[ & days due=max\{0,y-x-1\} \\ \hline \hline
13 & Jack &] 03/13, 03/17[ & 3\\ \hline
13 & George & ]03/06,  03/13[ & 6\\ \hline
13 &Karl & ] 03/13, 03/13[ & 0\\ \hline
\end{tabular}
\end{center}
\end{table}
Only 3+6+0=9 days will be accounted, which is obviously not enough.

The closed-open interval (respectively the open-closed one)  will 
give the following table \ref{Tab4}:
\begin{table}
\begin{center}
\caption{\label{Tab4}the closed-open interval's solution}
\vspace{2pt}
\begin{tabular}{|c|c|c|c|}\hline
Bed & Name & stay [x,y[ & days due=y-x \\ \hline \hline
13 & Jack &[03/13, 03/17[ & 4\\ \hline
13 & George & [03/06,  03/13[ & 7\\ \hline
13 & Karl & [03/13, 03/13[ & 0\\ \hline
\end{tabular}
\end{center}
\end{table}

The total due for bed 13, during the period beginning at 03/06 and 
ending at 03/17 is 4+7+0=11. If bed 13 is not
  used on 03/18, the day 03/17 is not perceived and all the day 03/13 
is due by Jack, Karl paying nothing.

We get three different solutions : 14, 9 and 13 days paid. None of 
these solutions
  gives the good result, that is 12 days.

\section{A new kind of interval}
  A granularity induces a sequence
of points on the physical time line, which are the meeting points of granules.
  Each point can be viewed as the representative of the granule that 
just follows it (in the sense of the arrow of Time).
In that sense, a granule is the set of all events
occurring during that period of time. This point of view is close to 
the natural meaning of a date, which is potentially
refinable until all events have been ordered (in a world where no 
synchronization occurs due to an ideal clock
  which is absolutely precise)\footnote{We are aware that this is not 
inside our physical  world where the Planck
time ($10^{-44}$ seconds) is the physical limit of precision. This 
clock is neither a time unit of a limit calendar, as
we proved it in \cite{Sch2002}.}. We first revisite Allen's 
qualitative relationships between a period and a granule with respect
to the granule, that is the result on a granule of the non-vacuous 
intersection between a period and a granule. Secondly, we revisite
chronologies and calendars in order to qualify a granule inside its 
own granularity. We end this section providing tools for denoting
how two periods can share a same granule.
\subsection{Allen's relations}
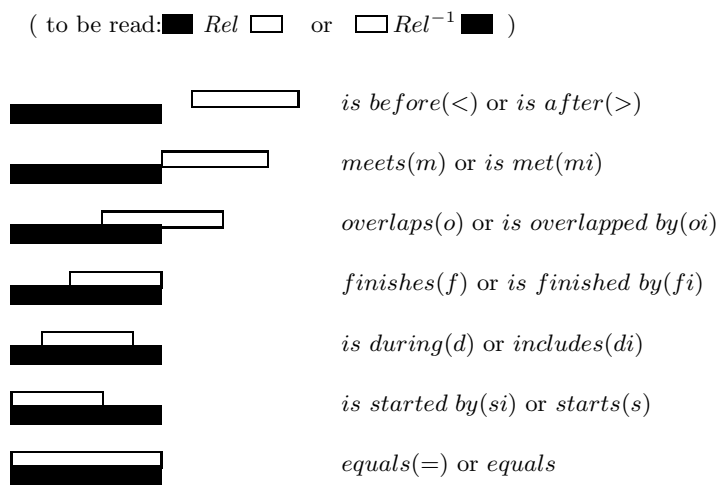
\begin{figure}
\setlength{\unitlength}{4mm}
\caption{\label{All13} Allen's atomic relations }
\begin{center}
\setlength{\unitlength}{0.4cm}
\begin{picture}(22,17)
\linethickness{0.25cm}
\multiput(0,0)(0,2){7}{\line(1,0){5}}
\put(5.05,15){\line(1,0){1}}
\put(6.4,14.75){{\em Rel}}
\put(10,14.75){or}
\put(0.6,14.75){( to be read: }
\put(16.5,14.75){)}
\put(12.7,14.75){$ Rel^{-1}$}
\put(15,15){\line(1,0){1}}

\thinlines
\put(8,14.75){\framebox(1,0.5)}
\put(11.5,14.75){\framebox(1,0.5)}

\put(0.05,0.25){\framebox(4.95,0.5)}
\put(11,0.15){$equals $(=) or $equals$}
\put(0.05,2.25){\framebox(3,0.5)}
\put(11,2.15){$is\  started\  by (si)$ or $ starts (s)$}
\put(1.05,4.25){\framebox(3,0.5)}
\put(11,4.15){$is\ during (d)$ or $includes (di)$}
\put(2,6.25){\framebox(3,0.5)}
\put(11,6.15){$finishes(f)$ or $is\ finished\ by(fi)$}
\put(3.05,8.25){\framebox(4,0.5)}
\put(11,8.15){$overlaps(o)$ or $is\ overlapped\ by(oi)$}
\put(5.05,10.25){\framebox(3.5,0.5)}
\put(11,10.15){$meets(m)$ or $is\  met(mi)$}
\put(6.05,12.25){\framebox(3.5,0.5)}
\put(11,12.15){$is\ before(<)$ or $is\ after(>)$}
\end{picture}
\end{center}
\end{figure}

The 13 possible relations between two symbolic intervals were set by 
Allen  \cite{All81} and we
show  them in Figure
  \ref{All13}.

In \cite{AlH85}, the relation {\em meet} is proved to be the 
generator of the 13 Allen relations,
and induces the notion of  Russel's point as an equivalence class.
Also in \cite{AlH85}, in order to avoid confusion between the span of 
time taken by an event and
  its mathematical representation,
  the term  {\it period} is preferred to the term {\it interval}, and 
the term {\it moment} is preferred to the term {\it point}
  but they are not themselves periods, not even very
short ones.
Two periods {\em meet} where there is neither stretch of time between 
them or stretch of time that they share.
Moments in time are ``places'' where periods meet. A moment is then 
an abstract object,
the nature of which is different from the interval nature.

A time period is the sort of thing that an event might occupy. Even a 
flash of lighting, although momenttlike
in many ways, must be a period because it contains a real physical 
event. Other things can happen at the same
time as the flash.

\subsection{Chronology revisited}
  In \cite{Sch2002}, has been noticed that there are two kinds of 
objects in a calendar, firstly there are {\em chronologies},
  associated to each unit,
  that
  allow to situate events and to measure periods of time, and secondly there are
{\em relationships} between chronologies,
  that allow to
situate and measure each event inside the best chronology for it and 
then to compare and to
synchronize all events in a same framework.
It is possible to measure the period of time taken by an event inside 
a chronology (an interval of
measure is given by the
number of occurrences of unit that is included in that period and the
  number of occurrences of unit that have a non vacuous intersection 
with that period.
But when one says that two occurrences of the unit {\em Month} do not 
have the same length,
it is supposed
implicitly that there is a way to measure or compare these two 
occurrences, that is with respect
to either a relationship with another chronology (for instance: {\em 
Day}) or a revealed time domain.

The path from one chronology into another one is a change of granularity.
  In recent works, calendars
  are considered as a particular case of the theory of granularity 
\cite{Hob85,Euz95,BWJ98}.
  These works deals with functions
that convert a measure or a mark from one chronology into another one 
when some constraints
of commensurability and of synchronization of the origins are achieved.

\cite{Sch2002} provides a categorical
view of these functions which are the elementary objects on which 
elementary operations,
called functors, are applied;
  that is operations
  well-known in usual arithmetics on the set of natural numbers like 
addition for concatenating
two calendars, or multiplication of a morphism by an ordinal for 
iterating a calendar.
In that framework,
  a chronology is defined as a couple $\langle \alpha , U \rangle $ 
where $\alpha$ is
  an ordinal such
  that  $0 < \alpha \leq \omega$, named its temporal domain, and U is its unit.
\begin{example}  $ \langle \omega , Day \rangle$  and $ \langle 
\omega , Hour \rangle$ are infinite chronologies;
   $ \langle 24 , Hour \rangle$   is a finite chronology.
\end{example}

Finite chronologies are very useful because we can iterate or
aggregate them for building new ones. Iteration allows building 
periodic chronologies. It is worth to notice that
addition (used to concatenate) and multiplication (used to iterate) 
on ordinals are not commutative (except in the finite case). But
the commutativity has to be strictly avoided as far as chronologies 
are concerned, so that, staying in the framework of ordinals can
help  not to use this property, that aims to erase the time dimension.

Inside a chronology, a unit is the thickness of the {\em now}. 
$\alpha$ is the number of units taken into account, often
called {\it  time-windowing function}, for example in \cite{WJS93}.
As such, it is the well-ordered set. The $\alpha$ elements of the 
ordinal $\alpha$ are its elements.
But neither $\alpha$ nor its  elements, that we yet call occurrences 
of unit (or units for short),
  are those which are thought when devising of the two natures of an ordinal.
These two natures of indivisibility or divisibility, depending of 
what Hobbs \cite{Hob85}
calls a change of granularity.

But nothing is said about ordinals as elements inside a chronology. 
We argue that the true nature of such an ordinal inside a chronology
is  {\em granule}.

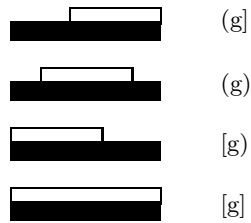
\begin{figure}
\setlength{\unitlength}{4mm}
\caption{\label{All} How a white period may intersect a black granule }
\begin{center}
\setlength{\unitlength}{0.4cm}
\begin{picture}(10,8)
\linethickness{0.25cm}
\multiput(0,0)(0,2){4}{\line(1,0){5}}

\thinlines

\put(0.05,0.25){\framebox(4.95,0.5)}
\put(7,0.0){[g]}
\put(0.05,2.25){\framebox(3,0.5)}
\put(7,2){[g)}
\put(1.05,4.25){\framebox(3,0.5)}
\put(7,4){(g)}
\put(2,6.25){\framebox(3,0.5)}
\put(7,6.15){(g]}
\end{picture}
\end{center}
\end{figure}

A granule is not a moment, but an atomic period. It is not possible 
to cut it (which is unacceptable for an atomic
object) but as an interval, it can be in one of the 13 atomic Allen's 
relations with any period. Restricted to the granule,
  there are only four possible relations\footnote{the notations will 
be defined in more detail in the following subsection.},
  that is only four ways  for a period to intersect a granule,
as shown in figure \ref{All}.

For instance, the relation between the intersection of period of the 
first worldwide war with year 1914 is
  (1914], with 1915 is [1915], with 1916 is [1916], with 1917 is 
[1917] and with 1918 is [1918).
The symbols [ and  ] have the same meaning as for usual intervals.
The two others symbols ``('' [resp.\ ``)'']  mean that the moment,
thought as an  ordered set is not wholly taken  but only a finishing 
[resp.\ a beginning] section of  it is taken.

Thus an endpoint of an interval can be either excluded, or included 
or partially included. We then have nine type  of
  intervals depending of the status of the two endpoints.
These types are only qualitative because nothing is said about
the part which is included. But this suffices to answer correctly to 
the following question: ``Knowing the
two worldwide war periods (14, 18) and (39, 45), what is the 
inter-war period?'' The answer is (18,39).

Granules have many of the properties of moments: if a period has 
moments at its ends,
then these granules are unique, and they uniquely define the period 
between them.
But granules also differ from moments in many ways, for instance they 
have distinct endpoints.

  Calendars, in the databases community, are defined upon the
discrete ordered set of chronons, which is
a  partition with finite intervals of the physical time line. A 
chronon, which is a finite non vacuous  interval of
\Rmath, is treated exactly as if a mathematical point. But it is not 
a mathematical point because it is very
large with respect to the Planck time, so that lots of sequential 
things may appear during this leap of time inside
  the system, but the system will work on them as if they where 
simultaneous. We will argue that they are neither a
moment nor a period even if, projected on the temporal linear and 
continuous line, they seem to be intervals as periods are,
so that periods are viewed as interval  of intervals. This explains 
why usual mathematical definitions of intervals, which
have to be either closed or open with respect to their ends, is not 
the right framework.

We suggest to adopt the denomination
of a {\em granule} for this kind of object.
We now recall the vocabulary of \cite{Sch2002}.
A chronology is potentially made for cover all the physical Time 
line, but the need for
  changing units brings to the definition of calendars, to go from a 
unit to a coarser or a finer one.
Hence  we set:
\begin{definition}[chronology]\

A {\em chronology} is a couple $\langle \alpha , U \rangle $  where 
$\alpha$ is  an ordinal such
  that  $0 < \alpha \leq \omega$, named its temporal domain, and U is its unit.
\end{definition}

\begin{definition}[atomic calendar]\

Let $\langle \alpha , U \rangle $ and $\langle \beta , V \rangle$ be 
two chronologies
  such that $\alpha \leq \beta \leq \omega $ and let  $f_{UV}$  a 
morphism from $\alpha$  into $\beta$.
The data $\langle U, V, f_{UV} \rangle$ defines a structure named an 
{\em atomic calendar}.

  If $\alpha=\beta=\omega$ then $\langle U, V, f_{UV} \rangle$ is an 
{\em atomic $\omega$-calendar}.
\end{definition}

  An atomic calendar has two commensurable units. A calendar is a 
directed acyclic graph where the set of nodes
  is the set of units, the  set of vertices is the set of morphisms 
such that, if $U$ and $V$ are
neighbors, $f_{UV}$ is the vertex between them.
\subsection{Granules}
Hence, we add a new element between point and interval in a calendar. 
It has to be noted than granule is the prime element of
the description, because it is defined inside a chronology. Points 
and intervals are dual objects with respect to morphisms.
As far as reality is no more concerned inside a calendar, it is 
natural to use points and intervals.
\begin{definition}[granule, point, interval]\

\begin{enumerate}
\item Let $\langle \alpha , U \rangle $ be a chronology.

  $ x \in \alpha$  is a {\em granule} of
the chronology  $\langle \alpha , U \rangle $.
\item Let $\langle \alpha , U \rangle $ and $\langle \beta , V 
\rangle$ be two chronologies
and  $\langle U, V, f_{UV} \rangle$ be an atomic calendar.

  $ x \in \alpha$  is an {\em interval} with respect to
the chronology  $\langle \beta , V \rangle $  .
\item  Let $\langle \gamma , X \rangle $ and $\langle \alpha , U 
\rangle$ be two chronologies
and  $\langle X, U, f_{XU} \rangle$ be an atomic calendar.

  $ x \in \alpha$  is a {\em point} with respect to
the chronology  $\langle \gamma , X \rangle $.
\end{enumerate}
\end{definition}

An interval inside a chronology has a mathematical meaning, but a 
period of the linear timeline, mapped
inside a chronology is not exactly an interval as far as its limits 
are concerned. It can be useful to say if the
entire moment is taken or not.
This is why we add a new type of endpoint for partially included 
end-moment that we note:

\begin{itemize}
\item[$\bullet$]  $(g,-$\quad for a left-end-granule not totally 
included inside the period
\item[$\bullet$] $ -,g)$\quad for a right-end-moment not totally 
included inside the period
\end{itemize}
Figure \ref{Gra} shows a representation of  $(g,-$.
\begin{figure}
\setlength{\unitlength}{4mm}
\caption{\label{Gra}  }
\begin{center}
\setlength{\unitlength}{0.4cm}
\begin{picture}(15,1)
\linethickness{0.25cm}
\put(0,0){\line(1,0){5}}

\thinlines

\put(2,0.25){\line(0,1){0.5}}
\put(2,0.25){\line(1,0){10}}
\put(2,0.75){\line(1,0){10}}
\put(14,0.15){(g,-}
\end{picture}
\end{center}
\end{figure}

We are now able to set that inter-war 14-18 is the interval (1914, 1918).

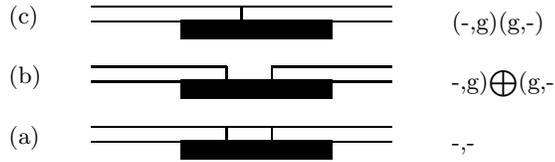
\begin{figure}
\setlength{\unitlength}{4mm}
\caption{\label{unio} How two periods share the same black end-granule g}
\begin{center}
\setlength{\unitlength}{0.4cm}
\begin{picture}(15,6)
\linethickness{0.25cm}
\multiput(5,0)(0,2){3}{\line(1,0){5}}

\thinlines
\put(-1,0.2){ (a)}
\put(8,0.25){\line(0,1){0.5}}
\put(6.5,0.25){\line(0,1){0.5}}
\put(2,0.25){\line(1,0){10}}
\put(2,0.75){\line(1,0){10}}
\put(14,0.0){-,-}
\put(-1,2.2){ (b)}
\put(8,2.25){\line(0,1){0.5}}
\put(6.5,2.25){\line(0,1){0.5}}
\put(2,2.25){\line(1,0){4.5}}
\put(2,2.75){\line(1,0){4.5}}
\put(8,2.25){\line(1,0){4}}
\put(8,2.75){\line(1,0){4}}
\put(14,2){-,g)$\bigoplus$(g,-}
\put(-1,4.2){ (c)}
\put(7,4.25){\line(0,1){0.5}}
\put(2,4.25){\line(1,0){10}}
\put(2,4.75){\line(1,0){10}}
\put(14,4){(-,g)(g,-)}
\end{picture}
\end{center}
\end{figure}

The following section will be devoted to the reasoning with such
interval, in order to get an answer to the hospital problem.
Before leaving this section, it is important to know how two periods 
share the same partial end-granule.
The three solutions, shown in Figure \ref{unio}, are given in the 
following picture and show three different kinds of union;
one with a non vacuous intersection, which is denoted by ``-,-'', one 
with the meet relations,
which is denoted by ``-,g)(g,-'' and the other with an interval gap 
in between which we name disjoint-union
and denote it by ``-,g)$\bigoplus$(g,-''.

We are now  ready to embark in the calculus.
\section{The calculus}
We first recall the calculus on usual interval, then we extend this 
calculus in the natural manner.
\subsection{The open/closed interval calculus}
As far as intervals are based on points, the general  definition of 
intervals inside a lattice $\langle L, \leq \rangle$ is

[a,b]= $ \{ x \in L |  a \leq x \leq b \}$

[a,b[= $ \{ x \in L |  a \leq x < b \}$

]a,b]= $ \{ x \in L |  a < x \leq b \}$

]a,b[= $ \{ x \in L |  a < x < b \}$

These four types are derived from the two ways that exist for typing 
each of the two endpoints:
included or excluded.
So we have

$\bullet$  $[x,-$ left included endpoint

$\bullet$  $] x,-$ left excluded endpoint

$\bullet$ $ -,x]$ right included endpoint

$\bullet$  $-,x[$ right excluded endpoint

Let us set $\sim$ the algebraic operator that converts an endpoint of 
an interval to the  corresponding endpoint of
the adjacent interval (that meets it in that endpoint), as shown in 
the table \ref{sim}:
\begin{table}
\begin{center}
\caption{\label{sim} Complementation}
\begin{tabular}{|c||c|c|c|c|}\hline
              &$ [x,-$ & $]x,-$ & $-,x]$  & $-,x[$ \\  \hline\hline
$\sim$ & $-,x[ $ & $-,x]$ & $]x,-$ & $[x,-$  \\ \hline
\end{tabular}
\end{center}
\end{table}
The binary operators union and intersection of two intervals that 
share a same endpoint
are given in the table \ref{ui}:

\begin{table}
\begin{center}
\caption{\label{ui} Union and Intersection}

\begin{tabular}{|c||c|c|c|c|}\hline
$\cup$     &$ [x,-$ & $]x,-$ & $-,x]$  & $-,x[$ \\ \hline \hline
$ [x,-$    &$ [x,-$ & $[x,-$ & $-,-$  & $-,-$ \\ \hline
$ ]x,-$    &$ [x,-$ & $]x,-$ & $-,-$  & $-,x[\cup]x,-$ \\ \hline
  $-,x]$    &$ -,-$  & $-,-$ & $-,x]$  & $-,x]$ \\ \hline
$-,x[$    &$ -,-$  & $-,x[\cup]x,-$ & $-,x]$  & $-,x[$ \\ \hline
\end{tabular}
\hspace*{5pt}
\begin{tabular}{|c||c|c|c|c|}\hline
$\cap$     &$ [x,-$ & $]x,-$ & $-,x]$  & $-,x[$ \\ \hline \hline
$ [x,-$    &$ [x,-$ & $]x,-$ & $[x]$  & $\emptyset$ \\ \hline
$ ]x,-$    &$ ]x,-$ & $]x,-$ & $\emptyset$ &$\emptyset$ \\ \hline
  $-,x]$    &$[x]$  &$\emptyset$ & $-,x]$  & $-,x[$ \\ \hline
$-,x[$    &$\emptyset$  &$\emptyset$ & $-,x[$  & $-,x[$ \\ \hline
\end{tabular}
\end{center}
\end{table}

  Mathematics works on ideal objects with null dimension: the points. 
A point is essentially atomic.
In the set of real number \Rmath, any point  is the limit of any 
infinite set of fitting together intervals which
contain it \cite[Weierstrass theorem]{Rud}. This is due to the 
completeness of \Rmath. The physical time-line is
usually modellized by a convex part of \Rmath.

\subsection{extended calculus}

In \Rmath, the closure of an interval is the closed interval with the 
same endpoints,
the opening of an interval is the open interval with the same endpoint.
That allow us to extend the $\sim$ operator
and these two topological operators (in \Rmath) as described in table 
\ref{extop}.

\begin{table}
\begin{center}
\caption{\label{extop} Extended $\sim$, opening and closure operations}

\begin{tabular}{|c|c|c|c|c|c|c|}\hline
              &$ [x,-$ & $(x,-$ &  $]x,-$ & $-,x]$  &  $-,x)$ & $-,x[$ 
\\  \hline\hline
$\sim$ & $-,x[ $ & $-,x)$ & $-,x]$ & $]x,-$& $(x,-$ & $[x,-$   \\ \hline
opening & $]x,- $ & $]x,-$ & $]x,-$ & $-,x[$& $-,x[$ & $-,x[$   \\ \hline
closure & $ [x,- $ & $ [x,-$ & $ [x,-$ & $-,x]$& $-,x]$ & $-,x]$   \\ \hline
\end{tabular}
\end{center}
\end{table}

The two operators are extended as shown in the table \ref{ExU}
where ``$\cup \cap$'' stands for ``$(x)$'' or ``$\emptyset$''
and
  ``$\cup \cup \cup $'' stands for ``$-,x) \bigoplus  (x,-$'' or ``$-, 
x)(x,-$'' or ``$-,-$''.\\

  \begin{table}
\begin{center}
\caption{\label{ExU} Extended union and Extended intersection}
\vspace{2pt}
\begin{tabular}{|c||c|c|c|c|c|c|}\hline
$\cup$     & $ [x,-$	 & $(x,-$ 
		 & $]x,-$  				 & $-,x]$ 
	 & $-,x)$  						& 
$-,x[$                   \\ \hline \hline
$ [x,-$    &  $ [x,-$ 	&$ [x,-$ 
		&  $[x,-$ 				& $-,-$ 
		& $-,-$ 
	& $-,-$                      \\ \hline
$ (x,-$    &  $ [x,-$ 	&$ (x,-$ 
	 &  $(x,-$ 				& $-,-$ 
	& $\cup\cup\cup$    & $-,x[\cup]x,-$                      \\ 
\hline
$ ]x,-$    &$ [x,-$		 & $(x,-$ 
		& $]x,-$   				& $-,-$ 
	 		& $-,x)\cup]x,-$ & $-,x[\cup (x,-$       \\ 
\hline
  $-,x]$    &  $ -,-$    & $-,-$        				& 
$-,-$   					& $-,x]$      & 
$-,x]$               & $-,x]$                      \\ \hline
$-,x)$    &    $ -,-$   &$\cup \cup \cup$& $-,x)\cup]x,-$    & $-,x]$ 
& $-,x)$ 							 & 
$-,x[$ \\ \hline
$-,x[$    &    $ -,-$   & $-,x[\cup (x,-$& $-,x[\cup]x,-$ & $-,x]$ & 
$-,x)$ 							 & $-,x)$ \\ 
\hline
\end{tabular}
\vspace{3pt}

\begin{tabular}{|c||c|c|c|c|c|c|}\hline
$\cap$      & $ [x,-$	 & $(x,-$ 
		 & $]x,-$  				 & $-,x]$ 
	 & $-,x)$  						& 
$-,x[$       \\ \hline \hline
$ [x,-$    &$ [x,-$     &    $(x,-$          & $]x,-$           & 
$[x]$     & [x)                    & $\emptyset$ \\ \hline
$ (x,-$    &$ (x,-$     &    $(x,-$          & $]x,-$           & 
$(x]$     & $\cup\cap$      & $\emptyset$ \\ \hline
$ ]x,-$    &$ ]x,-$ 		 &$ ]x,-$ 
			& $]x,-$ 		& $\emptyset$ & 
$\emptyset$ &$\emptyset$ \\ \hline
  $-,x]$    &$[x]$  			& $(x]$       	&$\emptyset$ 
		& $-,x]$ 				& $-,x)$ 
			 & $-,x[$ \\ \hline
$-,x)$    &$[x)$  			& $\cup\cap$  	&$\emptyset$ 
		& $-,x)$ 				& $-,x)$ 
			 & $-,x[$ \\ \hline
$-,x[$    &$\emptyset$ & 	$\emptyset$ &$\emptyset$ & $-,x[$ 
	&	$-,x[$ 						& 
$-,x[$ \\ \hline
\end{tabular}
\end{center}
\end{table}

Let us set, by the end, how these types are converted inside atomic 
calendars  hence,
(and hence by transitivity, inside calendars).

Let $\langle \alpha , U \rangle $ and $\langle \beta , V \rangle$ be 
two chronologies
  such that $\alpha \leq \beta \leq \omega $, let  $f_{UV}$ be a 
morphism from $\alpha$  into $\beta$
and $\langle U, V, f_{UV} \rangle$ the atomic calendar.  The reader 
will convince her/himself that the table \ref{chr}
is true.
  \begin{table}
\begin{center}
\caption{\label{chr}From one chronology to another one}
\vspace{2pt}
\begin{tabular}{|c||c|c|c|c|}\hline
              &$ [$ & $($ &  $]$ &$ )$ \\  \hline
$ \alpha \rightarrow  \beta$ & $[ $ & $($ or $ [$ & $]$ & $)$ or  $]$ \\ \hline
$ \alpha \leftarrow  \beta$  & $($ or $ [$& $( $ & $)$ or  $]$  & $)$\\ \hline
\end{tabular}
\end{center}
\end{table}
\section{Examples resolution}
\subsection{The Inter-war}
The period of the inter-war is the period between the two periods 
(1914,1918) and (1939,1945). These two intervals
  cover only a part of their endpoints. Hence, the period between has 
to cover the parts of the two endpoints
1918 and 1939 which are not covered by the two war periods and all 
the years between 1918 and 1945.
This period is then the intersection between the complementary of 
the two intervals ``-,1918)'' and  ``(1939,-''.
It is obtained first by using the $\sim$ operator on both endpoints 
``-,1918)'' and ``(1939,-'' which provides the
two intervals ``(1918,-'' and ``-,1939)'', secondly by the intersection of them
  which gives (1918,1939).
\subsection{The Hospital}
Jack's period in the hospital is  ``(03/06, 03/13)'',
that is two partial days (one beginning and one ending) and 5 full days.

  George's period is ``(03/13, 03/17)'', that is two partial days (one 
beginning and one ending) and 3 full days.

Karl's period is ``(03/13, 03/13)'', that is one partial day (middle).

The global  period is obtained by union of three periods with ``-, 
03/13)'',`` (03/13,03/13)'' and
``(03/13,-''.

The table gives ``-, 03/13) $\cup \cup \cup$ (03/13,03/13)  $\cup 
\cup \cup$ (03/13,-''.

There is {\it a priori}\/ $3 \times 3$=9 {\it scenarii}.
But it is obvious to any one that bed 13 cannot be shared by two patients
  at the same time and that there is a gap between two patients used 
this bed, so that,
according to the knowledge of the domain, there is thus only one {\it 
scenario}\/ which is
  ``(03/06, 03/13)$\bigoplus$ (03/13)$\bigoplus$ (03/13, 03/17)''.
The length of this period is the length of its
closure. The closure of a union is the union of the closure, hence 
this period is:
  ``[03/06, 03/13 ]$\cup$ [03/13]$\cup$ [03/13, 03/17]'' = ``[03/06,03/17]''.
Its length is 12 days.

All these informations may help the social security and the patients 
to have a fairer fee to pay!

\section{Conclusion}
In this paper, we introduced a new type of interval and the extension 
of operations on intervals to this new type
of interval based on the three different meanings of what it is 
usually called a granule inside a chronology, depending
on whether it is  viewed {\em  above}, {\em  under} or {\em inside} 
its chronology.
We showed on two examples how to use these new tools. We are going, 
in the REANIMATIC project,
to implement them and to provide translation functions between all 
kinds of chronologies inside a calendar.

Our proposition consists in the offer of a best approximation of what 
is due to the hospital,
{\it  i.e.}\ the possibility of signifying  if an endpoint day is 
totally or partially occupied by a patient.
Our solution gives the good solution and the number of users of a 
same beds per day.
Our proposition offers a good compromise between the hospital 
interest and patients'  one.
It will be possible to write that 03/13 has been shared by three patients,
but of course, without knowing proportionality because staying inside 
the same time unit.

This new type of interval is adequate not only for expressing time 
life period inside a chronology, but also for
translating Allen's relations between symbolic intervals inside a 
chronology. This concept of {\em partially
  included
end-granule} is very closed to the theory of granularity, inside 
which it would be used a lot as far as qualitative
reasoning is concerned.

\end{document}